\documentclass[final]{cvpr}

\usepackage{times}
\usepackage{epsfig}
\usepackage{graphicx}
\usepackage{amsmath}
\usepackage{amssymb}
\usepackage{bm}

\usepackage[pagebackref=true,breaklinks=true,colorlinks,bookmarks=false]{hyperref}

\def\cvprPaperID{5404} %
\def\confYear{CVPR 2021}

\begin{document}

\title{CanonPose: Self-Supervised Monocular 3D Human Pose Estimation in the Wild}

\author{Bastian Wandt\\
Leibniz University Hannover\\
Hannover, Germany\\
{\tt\small wandt@tnt.uni-hannover.de}
\and
Marco Rudolph\\
Leibniz University Hannover\\
Hannover, Germany\\
{\tt\small rudolph@tnt.uni-hannover.de}
\and
Petrissa Zell\\
Leibniz University Hannover\\
Hannover, Germany\\
{\tt\small zell@tnt.uni-hannover.de}
\and
Helge Rhodin\\
UBC Vancouver\\
Vancouver, Canada\\
{\tt\small helge.rhodin@ubc.ca}
\and
Bodo Rosenhahn\\
Leibniz University Hannover\\
Hannover, Germany\\
{\tt\small rosenhahn@tnt.uni-hannover.de}
}

\maketitle

\begin{abstract}
Human pose estimation from single images is a challenging problem in computer vision that requires large amounts of labeled training data to be solved accurately.
Unfortunately, for many human activities (\eg outdoor sports) such training data does not exist and is hard or even impossible to acquire with traditional motion capture systems.
We propose a self-supervised approach that learns a single image 3D pose estimator from unlabeled multi-view data.
To this end, we exploit multi-view consistency constraints to disentangle the observed 2D pose into the underlying 3D pose and camera rotation.
In contrast to most existing methods, we do not require calibrated cameras and can therefore learn from moving cameras.
Nevertheless, in the case of a static camera setup, we present an optional extension to include constant relative camera rotations over multiple views into our framework. 
Key to the success are new, unbiased reconstruction objectives that mix information across views and training samples.
The proposed approach is evaluated on two benchmark datasets (Human3.6M and MPII-INF-3DHP) and on the in-the-wild SkiPose dataset.

\end{abstract}

\section{Introduction}
\begin{figure}
	\centering
	\includegraphics[width=0.48\textwidth]{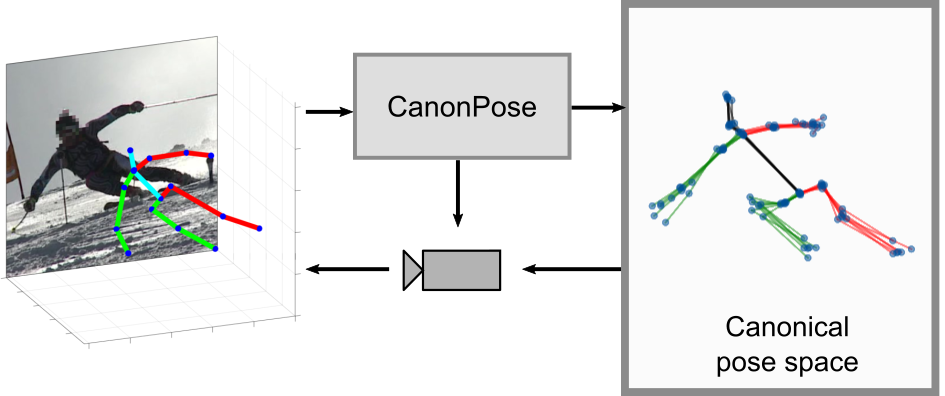}
	\caption{CanonPose learns a monocular 3D human pose estimator from multi-view self-supervision. By estimating 3D poses from different views in a canonical form together with the respective camera rotations we exploit multi-view consistency in the training data. Even for challenging outdoor datasets with moving cameras we achieve convincing 3D pose estimates from single images after training.}
	\label{fig:teaser}
\end{figure}
Human pose estimation from single images is an ongoing research topic in computer vision.
There exist a large amount of supervised deep learning solutions in the literature. %
These approaches achieve remarkable results in a supervised setting, \ie having 2D to 3D annotations, but heavily rely on a vast amount of available training data.
However, there are many activities a person can perform which are not present in common datasets.
For instance, human motions performed during outdoor and/or sports activities, \eg as shown in Fig.~\ref{fig:teaser}, are hard or even impossible to capture with a commercial motion capture systems.
Therefore, the acquisition of training data is a major practical challenge.
To this end, we propose a novel self-supervised training procedure that does not require any 2D or 3D annotations in the multi-view training dataset and works with uncalibrated cameras.
To acquire 2D joint predictions from images we use a 2D human joint estimator \cite{fang2017rmpe_alphapose} that is pretrained on a different dataset with only 2D joint annotations.
The only requirements for our method are at least two temporally synchronised cameras that observe the person of interest from different angles. 
No further knowledge about the scene, camera calibration and intrinsics is required.
Several related works consider a sparse set of 3D annotations \cite{rhodin2018learning,rhodin2019neural,Unsup3DPose}, unpaired 3D data \cite{WanRos2019a,Wang_2019_ICCV,Kundu_2020_CVPR}, or known camera positions \cite{rhodin2018learning,rhodin2019neural} to solve this problem.
However, such data rarely exists for outdoor settings with moving cameras.
To the best of our knowledge, there are only three competing methods \cite{chen2019unsupervised,kocabas2019epipolar,Iqbal_2020_CVPR} that apply to our setting.
They either require additional knowledge about the scene or observed person, such as scene geometry \cite{chen2019unsupervised} and bone lengths constraints \cite{Iqbal_2020_CVPR}, or sophisticated traditional computer vision algorithms that produce a pseudo ground truth pose \cite{kocabas2019epipolar}.

We propose a self-supervised training method which mixes outputs of multiple weight-sharing neural networks.
Fig.~\ref{fig:structure} shows our training pipeline when using two cameras.
Each individual network takes a single image as input and outputs a 3D pose in a canonical rotation, which gives our method its name \emph{CanonPose}.
This representation allows for the projection of all estimated 3D poses to any camera of the setup. 
Our approach splits into two stages.
The first stage predicts the 2D human pose from an image using a neural network pretrained on the MPII dataset \cite{mpii3Dhp2017}, in our case AlphaPose \cite{fang2017rmpe_alphapose,li2018crowdpose_alphapose}.
The second stage lifts these 2D detections to a 3D pose represented in a learned canonical coordinate system.
In a separate path it predicts the camera orientation to rotate the predicted 3D pose back into the camera coordinate system.
Combining the 3D pose from a first view with the rotation predicted from a second view, results in a rotated pose in the second camera coordinate system.
In other words, both 3D poses in the pose coordinate system should be equal and the predicted rotations project it back into the respective camera coordinate systems.
This enables the definition of a reprojection loss for each original and newly combined reprojection.
For static camera setups we propose an optional reprojection loss that is computed by mixing relative camera rotations between samples in a training batch.
Additionally, in contrast to existing self-supervised approaches, we also make use of the confidences that are typically provided by a 2D pose estimator for each predicted 2D joint by including them into the 2D input vector as well as into the reprojection loss formulation.

We evaluate our approach on the two benchmark datasets Human3.6M \cite{h36m_pami} and MPI-INF-3DHP \cite{mpii3Dhp2017} and set the new state of the art in several metrics for self-supervised 3D pose estimation.
Notably, this is without assuming any camera calibration or static cameras.
Our results are competitive to the fully supervised approach of Martinez et al. \cite{martinez_2017_3Dbaseline} which sets the baseline for single image pose estimation from 2D detections.
Additionally, we show results for the SkiPose \cite{sporri2016reasearch_skipose,rhodin2018learning} dataset.
This dataset represents all challenges that arise when activities are captured that cannot be performed in the restricted setting of a standard motion capture system.
It consists of outdoor scenes captured on a ski slope and includes fast motions, a large capture volume and pan-tilt-zoom cameras.

Summarizing, our contributions are:
\begin{itemize}
    \item We present CanonPose: a self-supervised approach to train a single image 3D pose estimator from unlabeled multi-view images by mixing poses across views.
    \item Our approach requires no prior knowledge about the scene, 3D skeleton or camera calibration.
    \item The proposed method directly employs multi-view images without any laborious pre-processing, such as camera calibration or multi-view geometry estimation.
    \item We integrate the confidences from the 2D joint estimator into the training pipeline.
\end{itemize}

\section{Related Work}
In this section we discuss recent 3D human pose estimation approaches by different types of supervision.

\paragraph{Full Supervision}
Recent supervised approaches rely on large datasets that contain millions of images with corresponding 3D pose annotations.
Li et al. \cite{Li2014} were the first to learn CNNs to directly regress a 3D pose from image input.
By integrating a structured learning framework into CNNs they later improved their work \cite{Li2015}.
Several others followed this end-to-end approach \cite{Tekin2016BMVC,Park2016,Du2016,VNect_SIGGRAPH2017,Pavlakos2017,lcrnet2017,Tome_2017_CVPR,sun2017compositional,OriNet2018,sun2018integral,zhou2019hemlets,Li_2020_CVPR,Xu_2020_CVPR,kocabas2019vibe}.
Typically, these end-to-end approaches perform exceptionally well on similar image data.
However, their ability to generalize to other scenes is limited.
Many works tackle this problem by cross dataset training or data augmentation.

There are other approaches that do not consider the image data directly but use a pretrained 2D joint detector \cite{chen20173d,Moreno_cvpr2017,fang2017learning,mpii3Dhp2017,pavlakos2018ordinal,XNect_SIGGRAPH2020}.
They benefit from training on large datasets that contain 2D annotations for many human activities in various scenes and are therefore agnostic to the image data.
Martinez et al. \cite{martinez_2017_3Dbaseline} directly train a neural network on 2D detections and 3D ground truth.
Due to its simplicity it can be trained quickly for many epochs leading to high accuracy and serves as a baseline for many following works.
The approach of \cite{martinez_2017_3Dbaseline} was extended by Hossain et al. \cite{Hossain2018} by employing a recurrent neural network for sequences of human poses.
While effective, the major downside of all supervised approaches is that they do not generalize well to unseen poses.
Therefore, their application to in-the-wild scenes is limited.

\paragraph{Weak Supervision}
Weakly supervised approaches only require a small set or even no annotated 2D to 3D correspondences.
An example for a commonly applied evaluation protocol for the Human3.6M dataset assumes that 3D annotations for one of the subjects of the training set are available.
A transfer learning approach is introduced by Mehta et al. \cite{mpii3Dhp2017} to allow for in-the-wild pose estimation of datasets where no training data is available.
This framework was later extended by Mehta et al. \cite{VNect_SIGGRAPH2017} to achieve real-time performance.
Rhodin et al. \cite{rhodin2018learning} use multi-view images and known camera positions to learn a 3D pose embedding in an unsupervised fashion.
The embedding facilitates the training with only a sparse set of 3D annotations.
This idea was adopted in other works \cite{rhodin2019neural,Unsup3DPose}.
Another approach is to employ unpaired 2D and 3D poses \cite{WanRos2019a,Wang_2019_ICCV,Kundu_2020_CVPR,zanfir20normalizing,chen20garnet,habekost20learning}.
Since these method learn distributions of plausible 3D poses and their properties they generalize better to unseen poses.
Although they are able to reconstruct out-of-distribution poses to a limited degree they struggle with completely unseen poses.

\paragraph{Self-supervised and Unsupervised Learning without 3D Ground Truth}
Recently, the interest in multi-view self-supervised and unsupervised 3D pose estimation is growing.
Our work also falls into this category.
Drover et al. \cite{drover2018can} propose an unsupervised approach to monocular human pose estimation.
They randomly project an estimated 3D pose back to 2D.
This 2D projection is then evaluated by a discriminator following adversarial training approaches.
Chen et al. \cite{chen2019unsupervised} extended \cite{drover2018can} with a cycle consistency loss that is computed by lifting the randomly projected 2D pose to 3D and inversing the previously defined random projection.
Although these two approaches are unsupervised, they integrate knowledge about the scene by constraining the camera rotation axis that is used for the random projection.
Rochette \etal \cite{rochette2019weakly} use a large amount of cameras from different viewing angles to achieve on par performance with a comparable fully supervised approach. 
However, due to the restriction to the camera setup the practical applicability is limited.
Kocabas et al. \cite{kocabas2019epipolar} propose a multi-view self-supervised approach which does not require any 3D supervision.
They apply traditional computer vision methods, namely epipolar geometry, to 2D pose predictions from multiple views to compute a pseudo ground truth which is then used to train the 3D lifting network.
Although this simple and effective straight-forward approach gives promising results, the laborious preprocessing step is very parameter sensitive and therefore does not generalize well. %
Moreover, mistakes due to wrongly estimated joints in the 2D prediction step result in a wrong pseudo ground truth.
Iqbal et al. \cite{Iqbal_2020_CVPR} tackle this problem by training an end-to-end network that refines the pre-trained 2D pose estimator during the self-supervised training.
Unfortunately, such approaches tend to easily overfit to a specific dataset.
For example, it could learn a background image for the training dataset which leads to exceptional performance on the specific dataset but does not generalize to other backgrounds.
This even happens in self-supervised settings.
Furthermore, Iqbal et al. \cite{Iqbal_2020_CVPR} employ a loss on normalized 3D bone lengths which is computed from the ground truth 3D poses of the Human3.6M training set. 

In contrast, our approach does not require knowledge about the scene and camera position or any anthropometric constraints.
Additionally, it is modular such that any 2D pose estimator can be used which makes it agnostic to the image data.
Even though our approach relaxes many constraints of the comparable works it still outperforms them in most experiments.

\section{Method}
\begin{figure}
	\centering
	\includegraphics[width=0.47\textwidth]{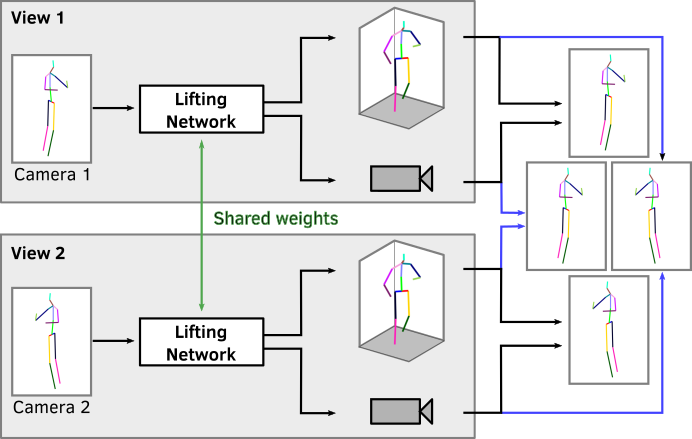}
	\caption{Network structure to learn single image 3D pose estimation from multi-view self-supervision. Each lifting network predicts a 3D pose and a camera rotation which is used to project the 3D pose back to 2D. Both networks observe the same 3D pose from different angles. We exploit this fact by applying the camera rotation to the respective other pose. This projects a predicted 3D pose into the other camera and gives an additional reprojection error. At inference time only one view (gray box) is applied.}
	\label{fig:structure}
\end{figure}
Our approach consists of two steps: first applying an off-the-shelf 2D joint detector to the input images, and second lifting these detections and the respective confidences for each joint to 3D.
The core idea of our approach is that 2D detections from one view can be projected to another view via a canonical pose space.
Fig.~\ref{fig:structure} shows our pipeline using two cameras.
For simplicity the network structure is shown for only two cameras.
If more cameras are available it is straight-forwardly extended.
A single neural network, the \textit{3D lifting network}, predicts the 3D pose $\bm{X}\in\mathbb{R}^{3\times j}$ with $j$ joints and a rotation $\bm{R}\in\mathbb{R}^{3\times 3}$ to rotate the pose to the camera coordinate system.
The pose is represented in a canonical pose coordinate system which is automatically learned during training.
Subsequently, the predicted 3D pose is rotated from the pose coordinate system to the camera coordinate system by the predicted rotation.
This separation into canonical human pose and camera rotation enables us to formulate various reprojection losses for self-supervision across views and samples.

\subsection{Reprojection}
\label{sec:reprojection}
Before a 2D pose is lifted to 3D it is normalized by centering it to the root joint and scaled by dividing it by its Frobenius norm.
This sidesteps the scale-depth ambiguity in monocular reconstruction.
In particular, the root centering gives a common rotation point for all 3D predictions.
For each view the predicted 3D pose is rotated into the camera coordinate system by $\bm{R}\bm{X}$.
$\bm{R}\in\mathbb{R}^{3\times 3}$ is a rotational matrix such that $\bm{R}\bm{R}^T=\bm{I}_3$ with $\bm{I}_3$ as the $3\times 3$ identity matrix and $det(\bm{R})=1$.
Since we assume weak perspective cameras, the projection to the camera plane is simply done by removing the depth coordinate, which is expressed as 
\begin{equation}
    \label{eqn:reprojection}
    \bm{W}_{\mathrm{rep}}=
    \begin{pmatrix}
    1 & 0 & 0 \\
    0 & 1 & 0
    \end{pmatrix}
    \bm{R}\bm{X},
\end{equation}
where $\bm{W}_{rep}\in\mathbb{R}^{2\times j}$ is called the reprojected 2D pose.
With $\bm{W}$ as the input 2D pose we define the \textit{scale-independent reprojection loss} as
\begin{equation}
    \label{eqn:rep_err}
    \mathcal{L}_{\mathrm{rep}} = \left\Vert \bm{W} - \frac{\bm{W}_{\mathrm{rep}}}{\|\bm{W}_{\mathrm{rep}}\|_F} \right\Vert_1,
\end{equation}
where $\| \cdot \|_1$ denotes the $L_1$ norm.
Since the global scale of the 3D pose is unknown and we consider weak perspective projections, scaling the reprojection $\bm{W}_{\mathrm{rep}}$ is essential.
Note that the input 2D pose $\bm{W}$ is already divided by its Frobenius norm in the preprocessing.
That means both, the input pose and the predicted pose, have the same scale.

To ensure that the network predicts a proper rotation, the matrix $\bm{R}$ is not predicted directly, but in axis-angle representation.
Let $(\theta)$ be a rotational angle and $\bm{\omega}=(\omega_1, \omega_2, \omega_3)$ denote a rotation axis.
With
\begin{equation}
    \bm{A}=
    \begin{pmatrix}
    0           & -\omega_3     & \omega_2  \\
    \omega_3    & 0             & -\omega_1 \\
    -\omega_2   & \omega_1      & 0
    \end{pmatrix}
\end{equation}
Rodrigues' formula is applied to obtain the rotation matrix
\begin{equation}
    \label{eqn:rodrigues}
    \bm{R}=\bm{I}_3 + (\sin \theta)\bm{A} + (1 - \cos \theta)\bm{A}^2
    .
\end{equation}

\subsection{View-consistency}
\label{sec:view_consistency}
A straight-forward way of ensuring view consistency would be to enforce a loss, such as $L_2$, between the canonical poses predicted by two views. 
In theory, that loss should be zero for the correct solution because the same person seen from two different views should have the same canonical pose.
In practice, however, this leads to the lifting network learning 3D poses that are view invariant but no longer in close correspondence to the input pose, preventing the network to converge to plausible solutions in our preliminary experiments.

The key insight to the proposed method is that rotations and poses from different views can be mixed to enforce the view consistency as a variant of the previously introduced reprojection objective. 
We mix the predicted camera and pose of two views, say view-1 and view-2, by rotating the predicted canonical 3D pose from the source view-1 to the target view-2 by using the rotation from view-2.
For two cameras as in Fig.~\ref{fig:structure} there exist four possible combinations of rotations and poses.
The same approach is easily extended to $m$ cameras which gives $m^2$ combinations.
During training time all possible combinations are reprojected to the respective cameras.
With this training scheme we enforce multi-view consistency without bias towards trivial solutions.
Note that the lifting network is only applied to a single frame at inference stage and does not need any other inputs.

\subsection{Confidences}
The output of most pretrained 2D joint estimators are 2D heatmaps where each entry indicates the confidence for the presence of the corresponding joint at the associated position in the image.
Commonly, the argmax or soft-argmax is computed and given as input to the following lifting network.
However, this gives an exact joint position independent of the confidence of the 2D detection.
That means uncertain predictions are processed in the same way as certain ones.
We circumvent this problem by two simple modifications.
First, we concatenate the maximum value of each heatmap, which is a surrogate to its confidence, to the 2D pose input vector to our lifting network.
Second, we modify the reprojection error in Eq.~\ref{eqn:rep_err} such that each difference between input and reprojected 2D is linearly weighted with its confidence by
\begin{equation}
    \mathcal{L}_{\mathrm{rep,c}}=\left\Vert\left(\bm{W} - \frac{\bm{W}_{\mathrm{rep}}}{\|\bm{W}_{\mathrm{rep}}\|_F}\right)\odot\bm{C}\right\Vert_1,
\end{equation}
where 
\begin{equation}
    \bm{C}=
    \begin{pmatrix}
        c_1 & c_2 & \dots & c_j \\
        c_1 & c_2 & \dots & c_j \\
    \end{pmatrix}
\end{equation}
with $c_i$ as the maximum value of the heatmap for joint $i$ and $\odot$ as the Hadamard product.

\subsection{Camera-consistency}
A reasonable assumption for many practical motion capture setups is that cameras are static during recording a sequence, \ie they do not change their position or orientation.
This is the case for the Human3.6M\footnote{In fact, camera angles change between subjects but not during a capture session with one subject.} and 3DHP dataset.
However, this assumption is not mandatory for our proposed method, but an enhancement for scenes with static cameras.
We will show the effect of this optional improvement in the experiments as well as the performance of our approach without it on the SkiPose dataset that contains moving cameras.

For a static camera setup all relative rotations between the cameras are equal.
An intuitive approach to enforce static cameras is to calculate an $L_2$-loss between the relative rotations over one batch of training samples.
However, a batch-wise loss leads to degraded solutions or had no effect if its weight was set to a low value.
This observation is similar to the findings regarding the canonical pose equality in Sec.~\ref{sec:view_consistency}.
For this reason we propose a similar mixing approach as in Sec.~\ref{sec:view_consistency}, now over estimates from different samples in one batch.
A relative rotation $\bm{R}_{1,2}$ using the rotation matrices $\bm{R}_1$ and $\bm{R}_2$ from view-1 to view-2 respectively, is defined by
\begin{equation}
    \bm{R}_{1,2}=\bm{R}_2 \bm{R}_1^T
    .
\end{equation}
Let $\bm{R}_{1,2}^{(s)}$ be the predicted relative rotation between view-1 and view-2 of sample $s$.
We then randomly permute these relative rotations in the batch and use them to reproject the canonical poses similar to Eq.~\ref{eqn:reprojection} 
\begin{equation}
    \bm{W}_{\mathrm{rep}}=
    \begin{pmatrix}
    1 & 0 & 0 \\
    0 & 1 & 0
    \end{pmatrix}
    \bm{R}_{1,2}^{(s)}\bm{R}_1^{(s')}\bm{X}^{(s')}
,
\end{equation}
where $\bm{R}_1^{(s')}$ and $\bm{X}^{(s')}$ are the rotation and estimated 3D pose in the current frame and $\bm{R}_{1,2}^{(s)}$ is the randomly assigned relative rotation from another sample in the batch\footnote{For the Human3.6M dataset we ensure that relative rotations are only changed in between subjects since camera positions vary between them.}.
The loss is calculated in the same way as the reprojection loss in Eq.~\ref{eqn:rep_err}.
Similar to Sec.~\ref{sec:view_consistency} this is easily extended to multiple cameras.
Again, we emphasize that this loss is optional to improve the results for the case of static cameras.
However, our method works without it.

\subsection{Network Architecture}
\begin{figure*}
	\centering
	\includegraphics[width=0.8\textwidth]{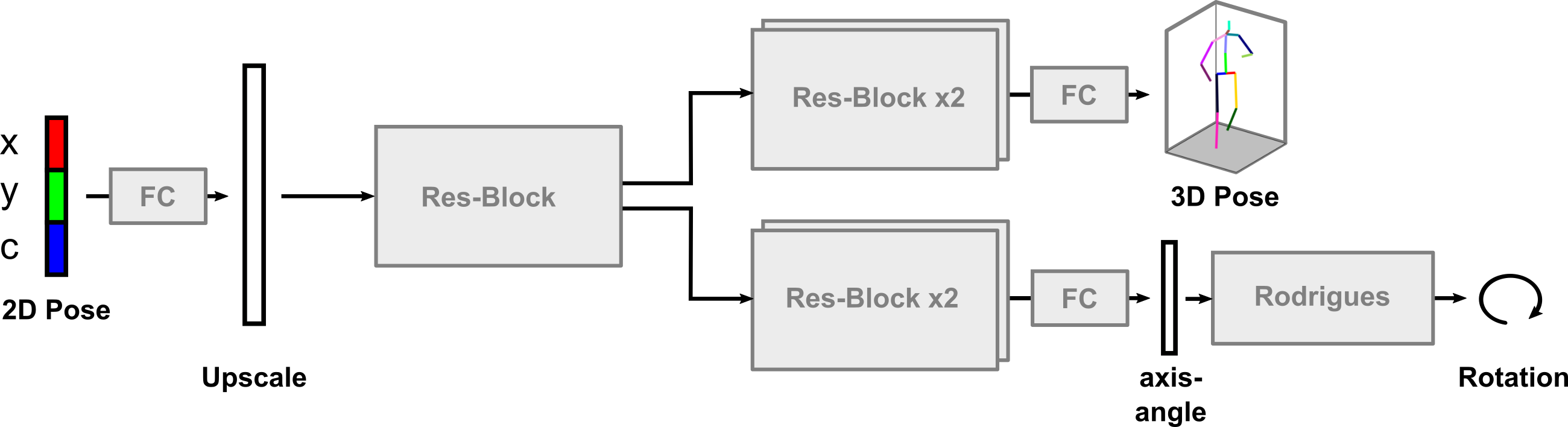}
	\caption{Network structure of the lifting network. The 2D input vector contains the $x$- and $y$-coordinates of the 2D pose and the confidence given by the 2D joint detector. It is upscaled using a fully connected layer with $1024$ neurons which then goes to a residual block. After that the network splits into two paths that predict the 3D pose in the canonical space and the camera rotation, respectively. Each of the paths has two consecutive residual blocks followed by a fully connected layer that downscales the features to the required size. The Rodrigues block implements Rodrigues formula (Eq.~\ref{eqn:rodrigues}) and has no trainable parameters.}
	\label{fig:lifting_network}
\end{figure*}
Fig.~\ref{fig:lifting_network} shows the architecture of our lifting network.
The input 2D pose vector is concatenated with a vector containing the confidences for each joint.
It is upscaled to $1024$ neurons by a one fully connected layer.
It is followed by a residual block consisting of fully connected layers with dimension $1024$.
Similar to \cite{WanRos2019a} the output is fed into two paths, each containing two consecutive residual blocks with identical architecture to the first block.
The 3D pose path directly outputs the 3D coordinates of the predicted pose in the pose coordinate system.
The camera path outputs a three-dimensional vector $\theta\bm{\omega}$ which is the axis angle representation.
The rotation matrix is computed using Rodrigues' formula as described in Sec.~\ref{sec:reprojection}.
The activation functions after each layer, except the two output layers, are leaky ReLU's with a negative slope of $0.01$.
We train the network for $100$ epochs using the Adam optimizer with an initial learning rate of $0.0001$ and weight decay at epochs $30$, $60$ and $90$, respectively.

\section{Experiments}
We perform experiments on the well-known benchmark datasets Human3.6M \cite{h36m_pami} and MPI-INF-3DHP \cite{mpii3Dhp2017}.
Additionally, we evaluate on the SkiPose dataset \cite{sporri2016reasearch_skipose,rhodin2018learning} to test the generalizability of our method to real world scenarios.
To conform with our setting of training a single image pose estimator with unlabeled images for a specific set of activities, we train one network for each dataset without using additional datasets.

\subsection{Metrics}
For the evaluation on Human3.6M there exist two standard protocols.
Both protocols calculate the \textit{mean per joint position error} (MPJPE), i.e. the mean euclidean distance between the reconstructed and the ground truth joint coordinates.
Since a multi-view self-supervised setting does not contain metric data, we adjust the scale of our predictions before calculating the MPJPE.
For a fair comparison with other works we compare to their scale adjusted predictions if they are available.
Protocol-I computes the MPJPE directly whereas Protocol-II first employs a rigid alignment between the poses.
Additional to the MPJPE one protocol for 3DHP calculates the \textit{Percentage of Correct Keypoints} (PCK).
As the name suggests it is the percentage of predicted joints that are within a distance of $150mm$ or lower to their corresponding ground truth joint.
\\\\
\textbf{Correct Poses Score (CPS)}
\\
For practical applications, such as motion analysis and prediction, the evaluation of the whole pose is a crucial prerequisite.
Even if a single joint of a pose is incorrect it can change downstream tasks significantly.
The formerly introduced metrics evaluate the quality of the prediction joint by joint.
However, they ignore the assignment of joints to poses and instead average over all joints in the test set.
Fig.~\ref{fig:cps} compares 3D pose estimates with their respective ground truths.
Each column shows two different reconstructions from the same pose.
The reconstructions in the top row have a lower PMPJPE compared to the bottom row.
However, the overall 3D poses appear better reconstructed in the bottom row.
In this section we present a simple yet powerful metric to evaluate such cases, the \textit{Correct Poses Score} (CPS).
A pose $\bm{W}$ is considered correct if for all joints $i$ the Euclidean distance is below a threshold value $\theta$.
Given a pose with joint positions $\bm{w}_i$ and predicted joint positions $\hat{\bm{w}}_i$ after rigid alignment, a correct pose is defined by
\begin{equation}
     \mathrm{CP}_\theta = 
      \begin{cases}
        1 & \|\bm{w}_i - \hat{\bm{w}}_i\|_2 < \theta \hspace{4mm} \forall i \in \{1, ..., j\} \\
        0 & else
      \end{cases}
      .
\end{equation}
Additional to the PMPJPE Fig.~\ref{fig:cps} shows the CP@180mm, which classifies the reconstructed poses into correct and incorrect.
The percentage of correct poses is calculated for the test dataset.
To be independent of the threshold, we calculate the area under curve for $\theta \in [0mm, 300mm]$ which defines the CPS.

\begin{figure}
	\centering
	\includegraphics[width=0.42\textwidth]{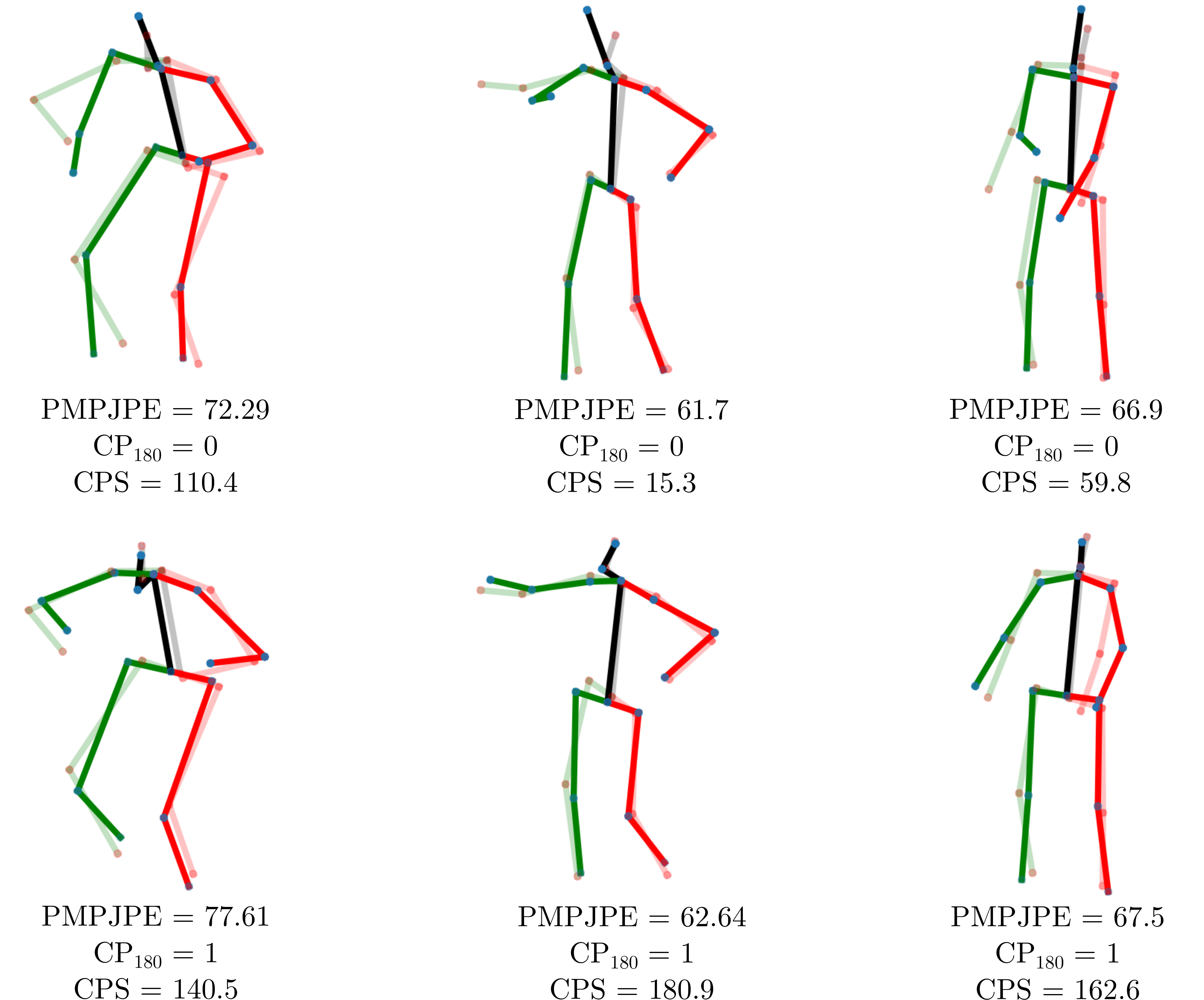}
	\caption{Comparison of PMPJPE and our CP-metric. Each column compares to different predicted 3D reconstructions with the same ground truth. While PMPJPE averages out high individual joint errors which are located in the right arm in the visualized case, CP indicates them. In this way, the correctness of the overall pose is evaluated. Note that for the calculation of the CPS we vary the threshold, which in these examples is $180mm$.}
	\label{fig:cps}
\end{figure}

\subsection{Skeleton Morphing}

We deploy an off-the-shelf detector AlphaPose~\cite{fang2017rmpe_alphapose} for retrieving the 2D human pose estimation required as input to our method.
The keypoint locations in the datasets used to train AlphaPose and other 2D pose estimation methods differ from the 3D skeleton of the test benchmarks.
For example, the root joint position is not in the middle of the hip joints and the relative position of the neck to the shoulders is different.
We circumvent this problem by training a \textit{2D skeleton morphing network}
that predicts the offset between the 2D pose from AlphaPose to the ground truth 2D pose in the dataset.
We train the morphing network on subject $1$ of each dataset with the given ground truth poses.
To not include these ground truth poses into our training, subject $1$ is excluded in all experiments. %
Thereby our data used for the self supervised training does not contain any 2D ground truth data, mimicking real application scenarios.
Note that the morphing network never sees any images and therefore is not able to learn domain specific image features.
In an experimental setting where the skeletal structure does not need to match a different skeleton this step is obsolete.
This is the case for most practical applications.

\subsection{Quantitative Evaluation on Human3.6M and 3DHP}
For the Human3.6M dataset, to keep it consistent with previous approaches, we follow standard protocols and evaluate only on every 64th frame.
However, with a sufficiently fast 2D pose estimator, which is the performance bottle neck of our complete pipeline, we can achieve real-time performance.
Table~\ref{tab:eval_h36m} shows the results of the proposed method compared to other state-of-the-art approaches.
We outperform every other comparable approach in terms of PMPJPE.
Note that we even achieve comparable performance to the fully supervised method of Martinez \etal \cite{martinez_2017_3Dbaseline} which has a lifting network with similar structure to ours.
Only one other self-supervised approach attains a lower MPJPE, however, by using additional information.
Our analysis revealed that although our pose structure is very accurate (which results in a low PMPJPE) the largest part of the error originates from a slight offset in the rotation.
For example, comparing frame $1$ from subject $9$ of the Human3.6M dataset to itself rotated by only $15^{\circ}$ around the longitudinal axis already results in an MPJPE of $67.7mm$.
Iqbal \etal \cite{Iqbal_2020_CVPR} still set the state of the art in terms of MPJPE.
However, they need bone length constraints which they directly compute from the ground truth 3D data of the training set.
Our approach does not require any predefined priors on the skeletal structure.
Using our static camera constraint (Ours+C) improves the MPJPE significantly.

\begin{table}[htp]
	\footnotesize
    \caption{Evaluation results for the Human3.6M dataset in $mm$. The bottom section, labeled with \textit{self}, shows methods that can solve our setting. Best results are marked in bold and second best in italic.}
	\centering
    \begin{tabular}{ l | l | c c c }
        Supervision & Method & MPJPE$\downarrow$ & PMPJPE$\downarrow$ \\
        \hline
        full                        & Martinez \cite{martinez_2017_3Dbaseline}  & 67.5 & 52.5\\
        \hline
        weak                        & Rhodin \cite{rhodin2018learning}          & 80.1  & 65.1 \\
                                    & Rhodin \cite{rhodin2018unsupervised}      & 122.6 & 98.2 \\
                                    & 3D interpreter \cite{3Dinterpreter2016}   & 98.4  & 88.6 \\
                                    & AIGN \cite{AIGN2017}                      & 97.2  & 79.0 \\
                                    & RepNet \cite{WanRos2019a}                 & 89.9  & 65.1 \\
                                    & HMR \cite{Kanazawa:CVPR:2018}             & -     & 66.5 \\
                                    & Wang \cite{Wang_2019_ICCV}                & 86.4  & 62.8 \\
                                    & Kolotouros \cite{SPIN:ICCV:2019}          & -     & 62.0 \\
                                    & Kundu \cite{Kundu_2020_CVPR}              & 85.8  & -    \\
                                    
        \hline
        \hline
        self
                                    & Chen \cite{chen2019unsupervised}               & -     & 68.0 \\
                                    & EpipolarPose \cite{kocabas2019epipolar}  & 76.6  & 67.5\\
                                    & Iqbal \cite{Iqbal_2020_CVPR}             & \textbf{69.1}  & 55.9 \\
        \hline
                                    & Ours                                      & 81.9  & \textbf{53.0}  \\
                                    & Ours + C                                  & \textit{74.3}  & \textbf{53.0} \\
    \end{tabular}
    \label{tab:eval_h36m}
\end{table}

\begin{figure*}
	\centering
	\includegraphics[width=0.98\textwidth]{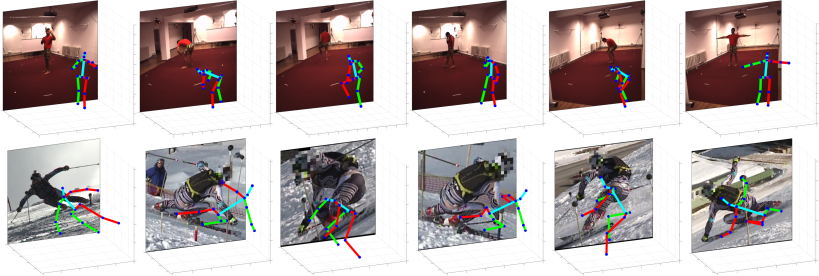}
	\caption{Qualitative results for the Human3.6M dataset (top) and for the challenging SkiPose dataset (bottom).}
	\label{fig:results}
\end{figure*}

Fig.~\ref{fig:cps} shows the CPS for our method compared to EpipolarPose \cite{kocabas2019epipolar}, which is the only comparable approach with publicly available code, and the 3D pose estimation baseline of Martinez et al. \cite{martinez_2017_3Dbaseline}.
On this metric, we outperform EpipolarPose by a large margin.
Note the high threshold of over $80mm$ that is required by \cite{kocabas2019epipolar} to achieve a CP above $1$\% compared to our threshold slightly below $50mm$.
As for the CPS metric, we are on par with the fully supervised approach of \cite{martinez_2017_3Dbaseline}.
Since their originally trained model is not publicly available anymore we retrained their model with their provided code to report the new CPS metric. The retrained model achieved a PMPJPE of $53.5$mm, which is slightly lower compared to their original number. The new model is used only for reporting CPS. 
Fig.~\ref{fig:results} shows qualitative results for the Human3.6M data set in the first row.

We also evaluate our approach on the 3DHP dataset \cite{mpii3Dhp2017} following the standard test protocols and metrics.
Table~\ref{tab:eval_3dhp} shows the results.
We outperform every other self-supervised approach.
In contrast to other approaches the proposed method does not require calibrated cameras\footnote{The configuration Ours+C only assumes that cameras are static during the sequence, which is a much weaker constraint.} or anthropometric constraints.
For the CPS metric we achieve a score of $134.2$.

\begin{table}[htp]
	\footnotesize
    \caption{Evaluation results for the 3DHP dataset. The bottom section, labeled with \textit{self}, shows methods that can solve our setting. Best results are marked in bold and second best in italic. MPJPE and PMPJPE are given in $mm$, PCK is in \%.}
	\centering
	\setlength\tabcolsep{4pt}
    \begin{tabular}{ l | l | c c c }
        Supervision & Method & MPJPE$\downarrow$ & PMPJPE$\downarrow$ & PCK$\uparrow$\\
        \hline
        weak                        & Rhodin \cite{rhodin2018learning}          & 121.8 & - & 72.7 \\
                                    & HMR \cite{Kanazawa:CVPR:2018}             & 169.5 & - & 59.6 \\
                                    & Habibie \cite{Habibie_2019_CVPR}          & -     & - & 70.4 \\
                                    & Kolotouros \cite{SPIN:ICCV:2019}          & 124.8 & - & 66.8 \\
                                    & Li \cite{Li20geometry}                    & -     & - & 74.1 \\
                                    & Kundu \cite{Kundu_2020_CVPR}              & 103.8 & - & 82.1 \\
        \hline
        \hline
        self
                                    & Chen \cite{chen2019unsupervised}               &   & 71.1 \\
                                    & EpipolarPose \cite{kocabas2019epipolar}  & 125.7 & - & 64.7 \\ 
                                    & Iqbal \cite{Iqbal_2020_CVPR}             & \textit{110.1}  & - & \textit{76.5}  \\
        \hline
                                    & Ours                                      & 119.2  & \textbf{68.7} & 69.0 \\
                                    & Ours + C                                  & \textbf{104.0} & \textit{70.3} & \textbf{77.0} \\
    \end{tabular}
    \label{tab:eval_3dhp}
\end{table}

\begin{figure}
	\centering
	\includegraphics[width=0.45\textwidth]{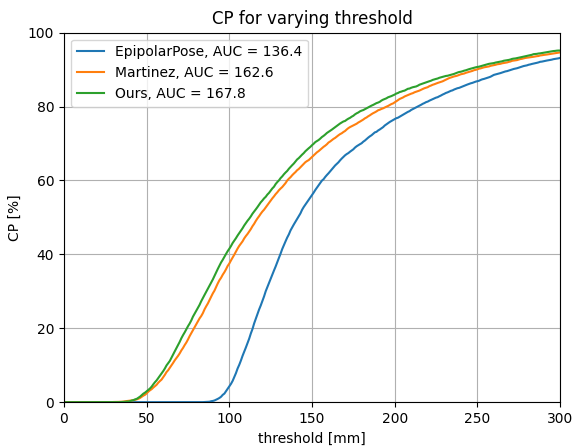}
	\caption{Comparison of CPS curves for distances from $1mm$ to $300mm$ with corresponding AUC for the Human3.6M dataset. A higher value means a better result, \ie the leftmost curve achieves the best result in terms of CP.}
	\label{fig:cps}
\end{figure}

\subsection{Moving cameras}
Our main motivation is to enable 3D human pose estimation in the wild by using a multi-view camera system with temporally synchronised cameras.
Moreover, the performed activity should be very challenging to capture and hard to simulate in a traditional motion capture studio.
That means a straight-forward activity domain transfer, \eg pretraining or combined training with a different dataset, is not reasonable. 
The SkiPose dataset \cite{sporri2016reasearch_skipose,rhodin2018learning} comprises all challenges of this motivation.
It features competitive alpine skiers performing giant slalom runs.
To record this dataset huge effort was taken to setup and calibrate the cameras and keep them in place after calibration.
Additionally, the cameras are rotating and zooming to keep the alpine skier in the field of view.
The proposed method can deal with all these difficulties since it does not require a calibrated or static setup and works with multiple synchronised cameras.
Since the camera setup is not static we cannot apply the relative rotation constraint here.
Table~\ref{tab:eval_skipose} shows our results in comparison to Rhodin \etal \cite{rhodin2018learning}.
Since they consider a (sparse-)supervised setting and known camera positions a direct comparison is not possible and only serves as a baseline.
Fig.~\ref{fig:results} shows qualitative results for the SkiPose dataset in the second row.

\begin{table}[h!tp]
	\footnotesize
    \caption{Evaluation results for the SkiPose dataset. The result for \cite{rhodin2018learning} was estimated from a bar plot in the paper. Since \cite{rhodin2018learning} considers a (sparse-)supervised setting and known camera position it is only shown as a baseline. MPJPE, PMPJPE and CPS are given in $mm$, PCK is in \%.}
	\centering
	\setlength\tabcolsep{4pt}
    \begin{tabular}{ l | l | c c c c }
        Supervision & Method                        & MPJPE$\downarrow$     & PMPJPE$\downarrow$    & PCK$\uparrow$   & CPS$\uparrow$\\
        \hline
        weak &
        Rhodin \cite{rhodin2018learning}                       & 85        & -         & -     & -\\
        \hline
        self &
        Ours                                                        & 128.1     & 89.6      & 67.1  & 108.7\\
    \end{tabular}
    \label{tab:eval_skipose}
\end{table}

\subsection{Ablation Studies}
To analyze our approach we perform a number of ablation studies.
First, to simulate a practical setting with limited resources, we reduced the number of cameras to train the model.
Table.~\ref{tab:ablations} shows the results for the training with only the first two or three cameras.
While the performance expectedly slightly drops due to the lower number of training samples and views our approach still produces good results which underlines its applicability in real world scenarios.
In a second experiment we show the impact of using the confidences from the 2D joint estimator as inputs to the network and for the calculation of the reprojection error.
They significantly impact the performance of our model and produce a gain of $19.4$mm in MPJPE and $11.2$mm in PMPJPE.
To prove that the proposed mixing of rotations and poses to achieve view- and camera-consistency is superior to simple equality constraints, we performed experiments with such equality constraints.
The results show that indeed our mixing approach is an essential part to make it work.
We also trained with ground truth 2D annotations to compute a lower bound for the proposed method.
\begin{table}[h!tp]
	\footnotesize
    \caption{Ablation studies on the Human3.6M dataset. All values are given in $mm$.}
	\centering
    \begin{tabular}{ l | c c c }
                                & MPJPE$\downarrow$     & PMPJPE$\downarrow$ & CPS$\uparrow$ \\
        \hline
        2 cams                  & 82.7 & 61.2 & 148.5 \\
        3 cams                  & 82.0 & 62.2 & 145.6 \\
        w/o confidences         & 95.6 & 65.0 & 142.5 \\
        ground truth 2D         & 65.9 & 51.4 & 187.1\\
        direct pose equality    & 554.3     & 360.8 & 0.0  \\
        direct camera equality  & 617.9     & 374.5 & 0.0  \\
        full (4 cams)           & 81.9 & 53.0 & 167.6 \\
        full+C (4 cams)         & 74.3 & 53.0 & 167.3 \\
    \end{tabular}
    \label{tab:ablations}
\end{table}

\subsection{Are We Learning a Canonical Pose Basis?}
\begin{figure}
	\centering
	\includegraphics[width=0.42\textwidth]{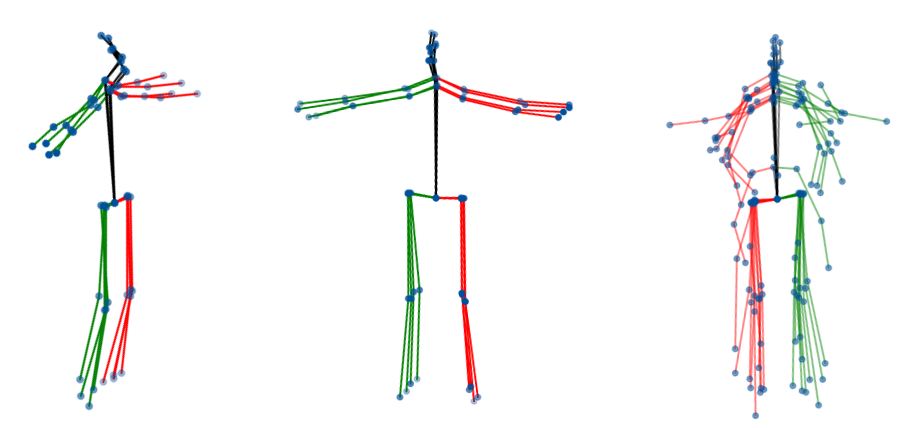}
	\caption{Visualization of the canonical pose space from the Human3.6M dataset. Left and middle: Canonical poses for the same 3D pose predicted from 4 different views. Right: 10 randomly sampled canonical poses. Our network automatically learns a disentanglement of a 2D pose into 3D and a camera rotation.}
	\label{fig:canonical1}
\end{figure}
Finally, we evaluate the claim that we learn a canonical pose basis.
To visualize the disentanglement for different 3D poses Fig.~\ref{fig:canonical1} shows a visualization of reconstructed 3D poses in the canonical basis obtained from 4 views on the left and in the middle.
The right image shows $10$ randomly picked reconstructions in the canonical space.
Although the similarity of the poses is not enforced directly as described in Sec.~\ref{sec:view_consistency} the poses are similarly oriented in the canonical space.
In particular, the hip joints are aligned which leads to a similar alignment of the upper body.
The standard deviation for the hip joints of the canonical poses from the test set of Human3.6M are $7.9mm$ and $7.7mm$ for the right and left hip, respectively.
This underlines that pose and rotation are disentangled plausibly by our network.

\section{Conclusion}
We present CanonPose, a neural network trained for single image 3D human pose estimation from multi-view data without 2D or 3D annotations.
Given a pretrained 2D human pose estimator we exploit multi-view consistency to automatically decompose a 2D observation into a canonical 3D pose and a camera rotation that is used to reproject it back to the observation after mixing.
Since our approach does not require either 2D nor 3D annotations for the multi-view data it is practically applicable to many in-the-wild scenarios, including outdoor scenes with moving cameras.
We not only achieve state-of-the-art results on benchmark datasets with less prerequisites compared to other approaches, but also show promising results on challenging outdoor scenes.

{\small
\bibliographystyle{ieee_fullname}
\bibliography{bibliography}
}

\end{document}